\documentclass[letterpaper]{article} 
\usepackage{aaai25}  
\usepackage{times}  
\usepackage{helvet}  
\usepackage{courier}  
\usepackage[hyphens]{url}  
\usepackage{graphicx} 
\usepackage{natbib}  
\usepackage{caption} 
\usepackage{algorithm}
\usepackage{algorithmic}

\usepackage{newfloat}
\usepackage{listings}
\DeclareCaptionStyle{ruled}{labelfont=normalfont,labelsep=colon,strut=off} 
\floatstyle{ruled}
\newfloat{listing}{tb}{lst}{}
\floatname{listing}{Listing}
\usepackage{booktabs} 
\usepackage{multirow} 
\usepackage{amsmath}
\usepackage{subcaption}
\usepackage{tcolorbox}
\usepackage{amssymb}
\usepackage{enumitem}

\definecolor{lightred}{rgb}{1.0,0.3,0.3}
\definecolor{lightblue}{rgb}{0.3,0.3,1.0}

\title{EDGE: Unknown-aware Multi-label Learning by \\ Energy Distribution Gap Expansion}
\author{
    Yuchen Sun\textsuperscript{\rm 1,2}, Qianqian Xu\textsuperscript{\rm 1}\thanks{Corresponding Author}, Zitai Wang\textsuperscript{\rm 1}, Zhiyong Yang\textsuperscript{\rm 2}, Junwei He\textsuperscript{\rm 1,2}
}
\affiliations{
    \textsuperscript{\rm 1}Key Laboratory of Intelligent Information Processing, Institute of Computing Technology, CAS, Beijing, China\\
    \textsuperscript{\rm 2}School of Computer Science and Technology, University of Chinese Academy of Sciences, Beijing, China\\
    \{sunyuchen22s, xuqianqian, wangzitai\}@ict.ac.cn, yangzhiyong21@ucas.ac.cn, hejunwei22s@ict.ac.cn
}

\usepackage{bibentry}

\begin{document}

\maketitle

\begin{abstract}
Multi-label Out-Of-Distribution (OOD) detection aims to discriminate the OOD samples from the multi-label In-Distribution (ID) ones. Compared with its multiclass counterpart, it is crucial to model the joint information among classes. To this end, JointEnergy, which is a representative multi-label OOD inference criterion, summarizes the logits of all the classes. However, we find that JointEnergy can produce an imbalance problem in OOD detection, especially when the model lacks enough discrimination ability. Specifically, we find that the samples only related to minority classes tend to be classified as OOD samples due to the ambiguous energy decision boundary. Besides, imbalanced multi-label learning methods, originally designed for ID ones, would not be suitable for OOD detection scenarios, even producing a serious negative transfer effect. In this paper, we resort to auxiliary outlier exposure (OE) and propose an unknown-aware multi-label learning framework to reshape the uncertainty energy space layout. In this framework, the energy score is separately optimized for tail ID samples and unknown samples, and the energy distribution gap between them is expanded, such that the tail ID samples can have a significantly larger energy score than the OOD ones. What's more, a simple yet effective measure is designed to select more informative OE datasets. Finally, comprehensive experimental results on multiple multi-label and OOD datasets reveal the effectiveness of the proposed method. 
\end{abstract}

\begin{figure}[t]
    \centering
    \begin{subfigure}{0.49\linewidth}
        \includegraphics[width=1.0\linewidth]{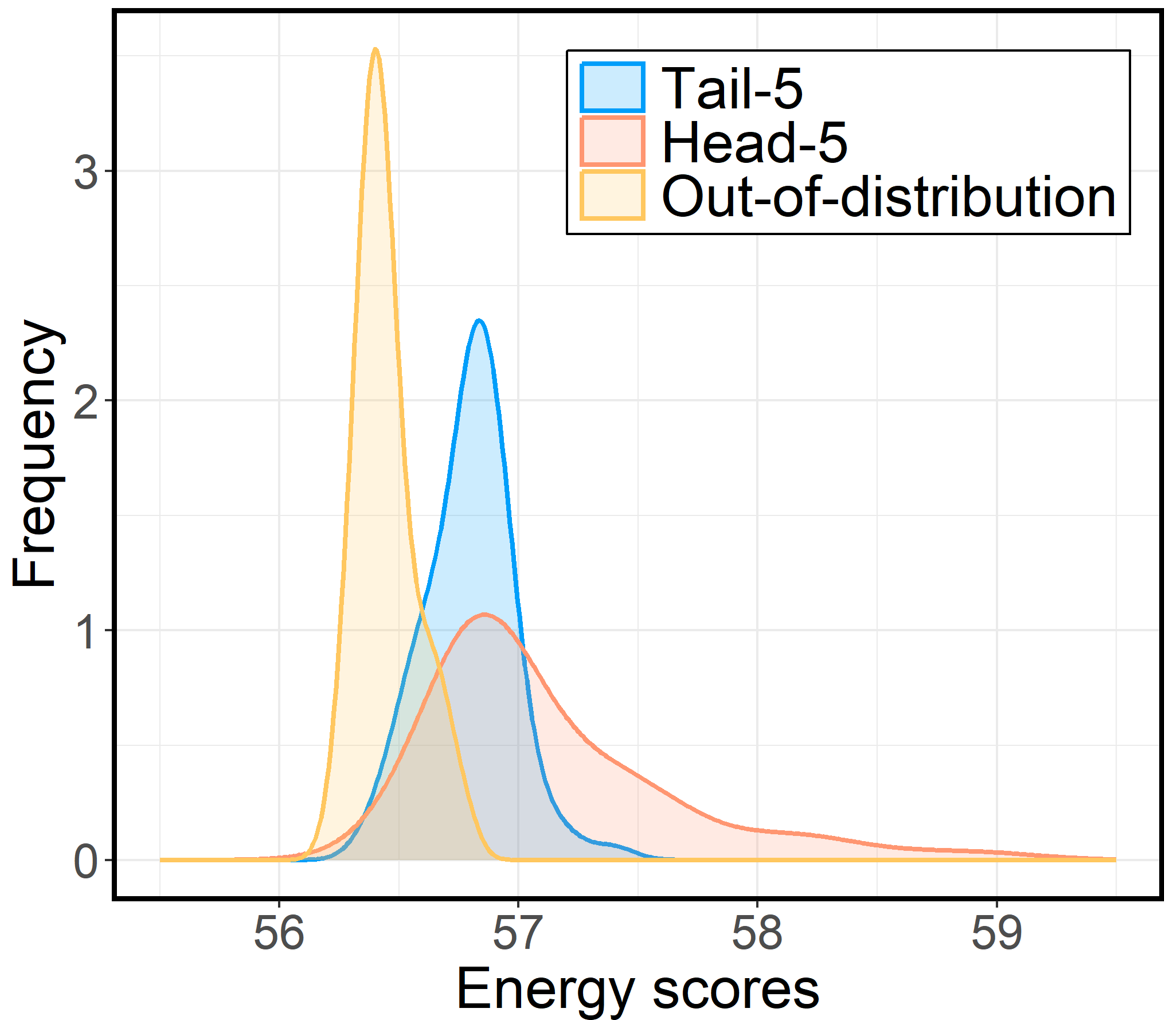}
        \caption{Distribution of baseline}
        \label{fig:normal_distribution}
    \end{subfigure}
    \hfill
    \begin{subfigure}{0.49\linewidth}
        \includegraphics[width=1.0\linewidth]{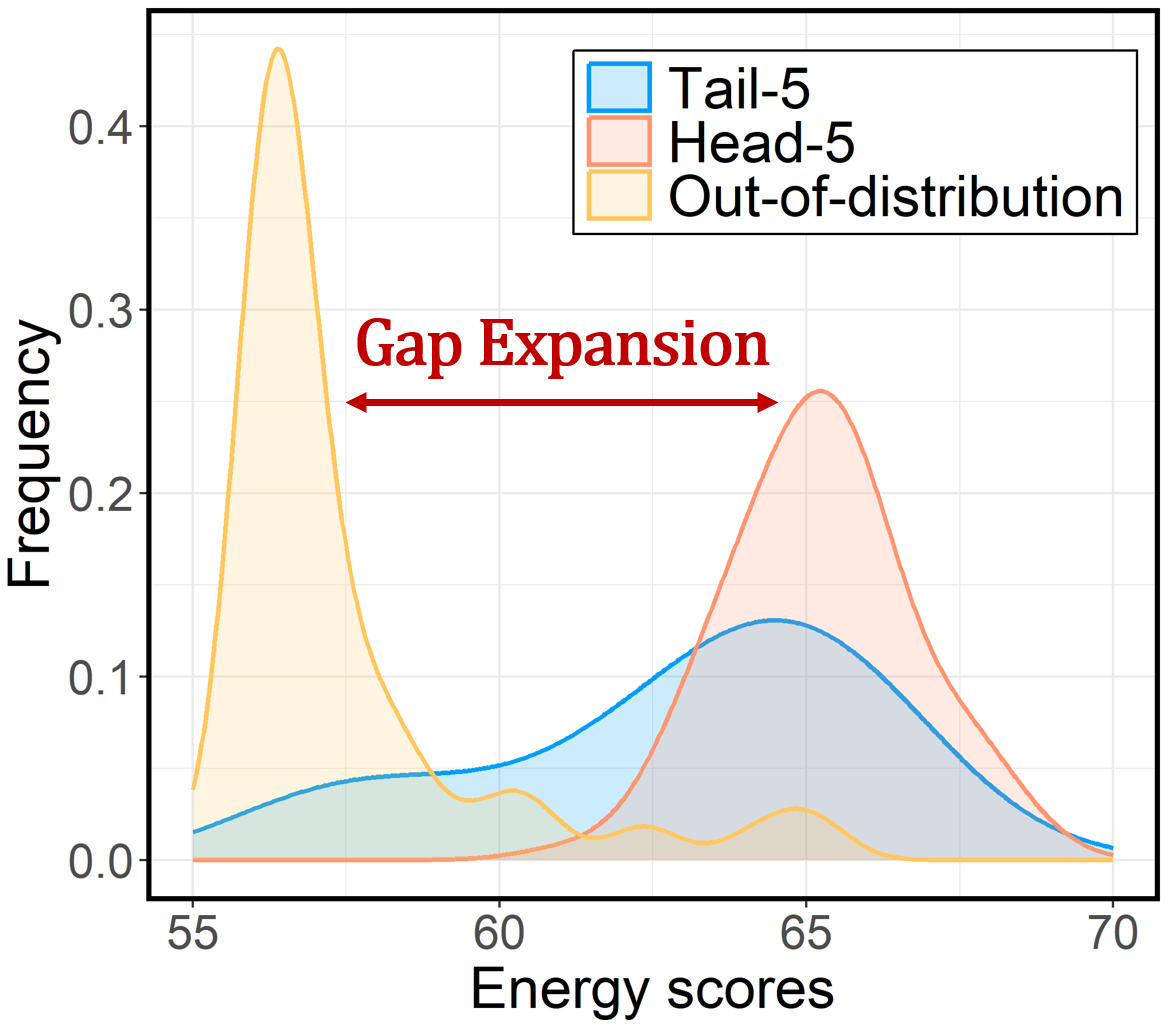}
        \caption{Distribution of EDGE}
        \label{fig:joint_distribution}
    \end{subfigure}
    \begin{subfigure}{0.49\linewidth}
        \includegraphics[width=1.0\linewidth]{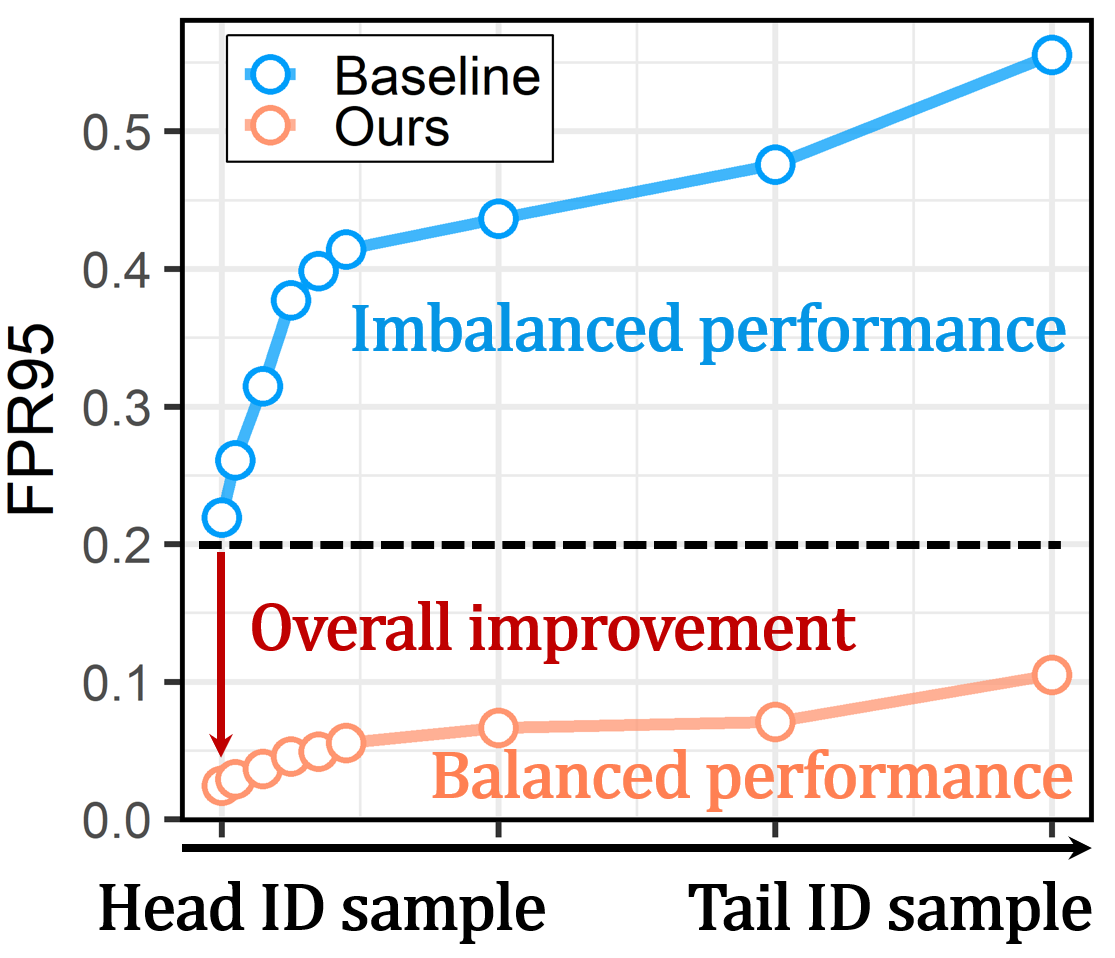}
        \caption{FPR95 comparison result}
        \label{fig:nus-wide_fpr95}
    \end{subfigure}
    \hfill
    \begin{subfigure}{0.49\linewidth}
        \includegraphics[width=1.0\linewidth]{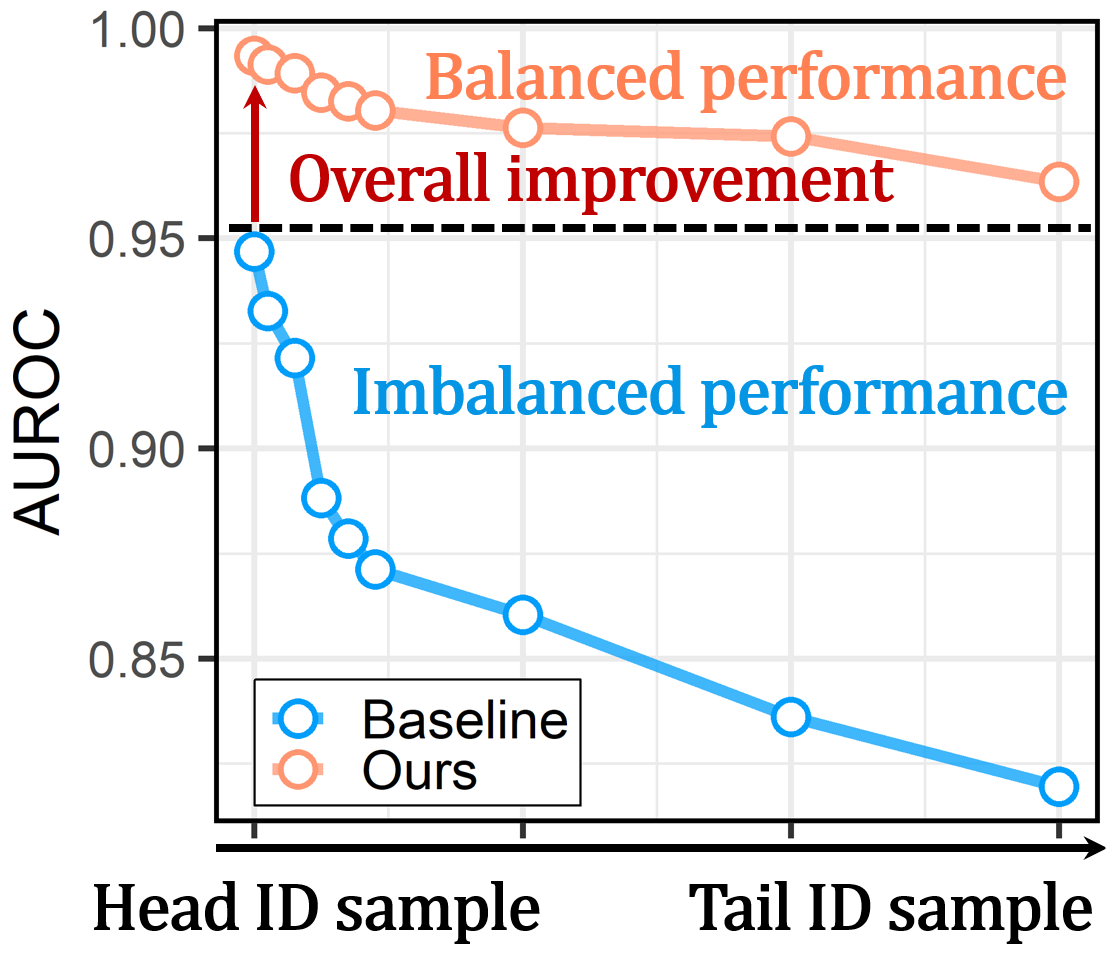}
        \caption{AUROC comparison result}
        \label{fig:nus-wide_auroc}
    \end{subfigure}
    \caption{Distribution and performance results with \textcolor[rgb]{ 0,  .62,  .98}{Binary Cross-Entropy loss (BCE)} training loss and \textcolor[rgb]{ 1,  .588,  .443}{our framework} on Places50 dataset. (a) shows the small gap between known and unknown domains under basic multi-label models, and (b) indicates the improvement of our method. The last two figures show the (c) FPR95 and (d) AUROC imbalanced and balanced results on different in-distribution subsets, where the proportion of in-distribution samples that only relate to minority classes increases with the x-axis value increases.}
    \label{fig:motivation_tendency}
    \vspace{-4mm} 
\end{figure}

\section{Introduction}

Traditional classification methods mainly focus on In-Distribution (ID) scenarios, where the training set and the test set share the same data distribution. However, in practical applications, it is unavoidable that some test samples are Out Of the Distribution (OOD) of the training ones \cite{DBLP:journals/corr/abs-2110-11334}. In this case, an ID model will classify the OOD samples into one of the ID classes due to the notorious overconfidence issue. Such wrong predictions seriously degenerate the model performance. Given this, OOD detection has attracted rising attention in recent years, intending to discriminate the OOD samples from the ID ones \cite{DBLP:conf/cvpr/HuangL21, DBLP:conf/nips/WangX00CH22, DBLP:conf/iclr/DjurisicBAL23}

In this direction, most prior art focuses on scenarios where the ID samples only have a single ground-truth label. Nevertheless, samples are inherently associated with multiple labels in many real-world applications such as autonomous driving \cite{DBLP:journals/ijcv/MaoSWL23} and medical image \cite{DBLP:journals/pami/ZhouLCYY23}. For multi-label OOD detection, one intuitive strategy is to adapt the off-the-shelf multiclass algorithms \cite{DBLP:conf/iclr/DuWCL22, DBLP:conf/icml/WangZZZLSW22}. However, Wang \textit{et al.} \cite{DBLP:conf/nips/WangLBL21}, pioneered in multi-label OOD detection, point out that simply using the largest model output (\textit{i.e.}, MaxLogit \cite{DBLP:conf/icml/HendrycksBMZKMS22}, Max) might be sub-optimal. Considering this, it is crucial to jointly model the information across different labels for multi-label tasks. To fix this issue, they suggest summing the energy score of all labels as the inference criteria (dubbed JointEnergy), where the energy score is proportional to the logit.

Although JointEnergy is effective for OOD inference, we are surprised to find that it induces a new imbalanced problem. As shown in Figure \ref{fig:motivation_tendency} (a), the energy gap between ID samples and OOD samples is relatively small, especially for the tail ID samples. Notably, in multi-label learning, \textit{tail} samples refer to instances whose relevant classes are all minority ones. In contrast, head samples are relevant to at least one majority class, where minority/majority indicates the class is associated with many/few samples, respectively. The intuition behind this problem is that traditional multi-label optimization methods (like binary cross-entropy loss) treat each class of samples equally while somewhat ignoring the focus on minority classes. Consequently, head samples tend to have large logits and thus are more likely to be recognized as ID samples. By contrast, some hard samples generally have low logits and are misclassified as OOD ones. Due to such a distribution overlap, the OOD performance degenerates and becomes significantly imbalanced across head samples and tail samples. As the blue line shown in Figure \ref{fig:motivation_tendency} (c) \& (d), we rank the classes according to the number of associated samples from largest to smallest. When the x-axis values increase, we remove all samples that are associated with the corresponding head categories and observe an obvious degradation in the FPR95 and AUROC for the remaining samples. 

\begin{figure*}[t]
  \centering
  \includegraphics[width=\linewidth]{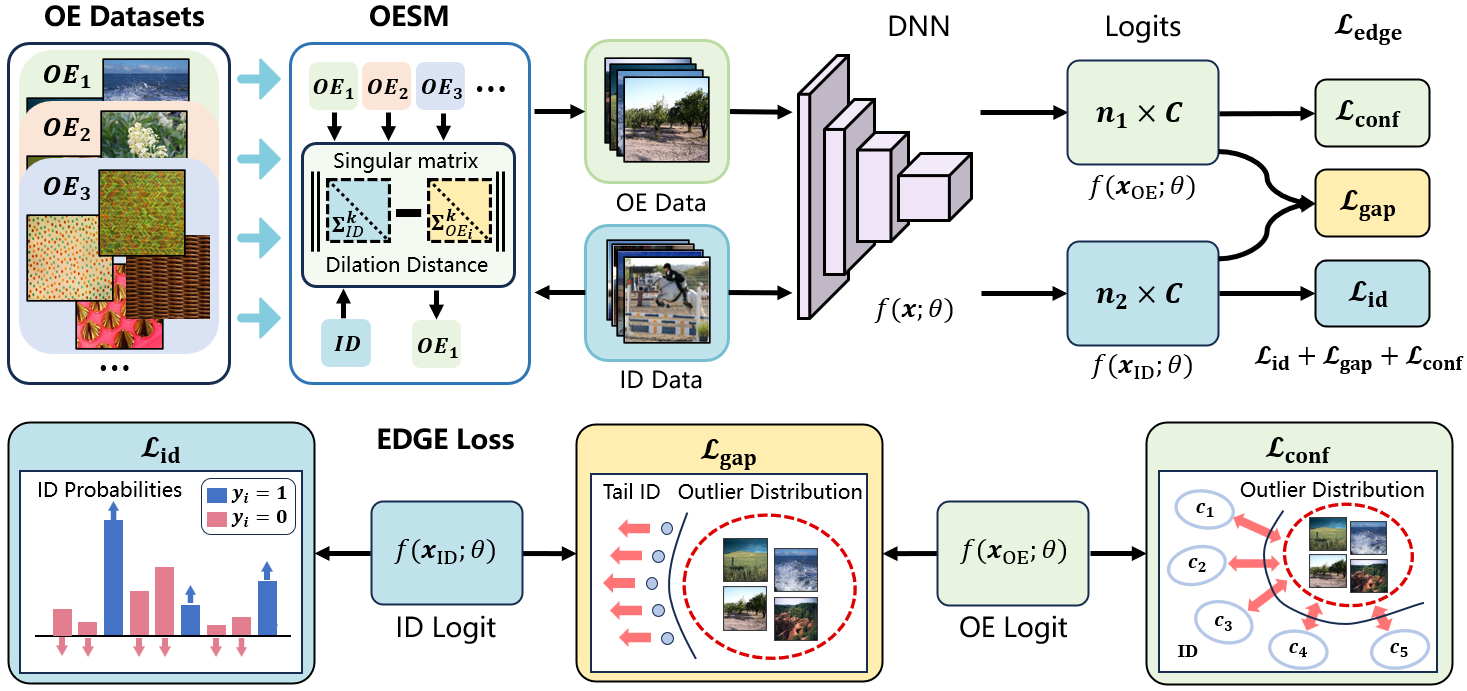}
   \caption{Overview of EDGE learning framework. This process could be divided into two parts, \textit{i.e.}, auxiliary OE data selection, and EDGE loss optimization. Firstly, the candidate OE datasets are fed into the OE Selection Module (OESM, shown in Fig. \ref{fig:osm}). Then, the deep model is provided with both in-distribution data and an informative auxiliary dataset to conduct unknown-aware multi-label learning with the two-part EDGE loss.}
   \label{fig:overview}
   \vspace{-4mm}
\end{figure*}

To tackle this challenge, we propose a simple yet effective unknown-aware multi-label learning framework named \underline{E}nergy \underline{D}istribution \underline{G}ap \underline{E}xpansion \textbf{(EDGE)}. Specifically, this method could be divided into three parts. Firstly, we enhance the generalization of models by learning known distribution. Then, we adopt the Outlier Exposure (OE) strategy, \textit{i.e.}, introducing auxiliary OOD data, to lower the logit of OOD samples and separate them from known classes. However, merely optimizing distribution space from the perspective of logit might suffer from a trivial solution. Thus thirdly, we further expand the positive energy gap between tail ID and OOD samples to establish a reliable energy boundary (see Figure \ref{fig:motivation_tendency} (b)). Compared with post-hoc methods, this framework could achieve a balanced performance, as the orange line shown in Figure \ref{fig:motivation_tendency} (c) \& (d). To sum up, the whole learning framework is shown in Figure \ref{fig:overview}, and the main contributions of this paper are three-fold:

\begin{itemize}
    \item  To the best of our knowledge, we are the pioneer in discovering the induced imbalance issue existing in the energy-based method for multi-label OOD detection. 
    \item A novel unknown-aware multi-label learning framework \textbf{EDGE} is proposed to reshape the uncertainty energy space of ID and OOD data, as well as a feature-based pre-processing module to guide informative outlier exposure data selection. 
    \item We conduct comprehensive experiments on three large-scale real-world datasets and various OOD datasets, where the test results consistently show the superiority of the EDGE framework.
\end{itemize}

\section{Related Work}
\label{sec:related work}

\subsection{Multi-label Classification}

Multi-label classification (MLC) focuses on the instances simultaneously assigned with multiple labels \cite{DBLP:journals/corr/abs-2407-06709, DBLP:journals/pami/LiuWST22, DBLP:conf/ijcai/HuangXWFY20}. This topic mainly falls into two categories: \textbf{problem transformation} and \textbf{algorithm adaptation}. For the former, most studies tend to simplify MLC into multi-class problems \cite{DBLP:conf/icml/JerniteCS17, DBLP:conf/nips/X19}. In contrast, algorithm adaptation leverages traditional machine-learning methods to model data \cite{DBLP:journals/corr/GongJLTI13, DBLP:conf/cvpr/WangYMHHX16,DBLP:conf/cvpr/ChenWWG19}. Unlike traditional MLC work, this paper is dedicated to exploring OOD detection in multi-label learning, which is still an underexplored problem. 

\subsection{Out-of-distribution Detection}

The study toward OOD detection is first suggested in \cite{DBLP:conf/cvpr/NguyenYC15}, and now mainly has two directions, \textit{i.e.}, \textbf{inference detection}, and \textbf{outlier exposure}. Representatively, \textbf{inference-based approaches} leverage informative outputs, like features and logits, to detect OOD inputs. For example, Energy Score \cite{DBLP:conf/nips/LiuWOL20} function identifies OOD samples from the view of energy; MSP \cite{DBLP:conf/iclr/HendrycksG17} detects OOD samples by maximum predicted softmax probability; and ODIN \cite{DBLP:conf/iclr/LiangLS18} pre-processes the inputs and rescales the logits with temperature. Recently, a few efficient methods like ViM \cite{DBLP:conf/cvpr/Wang0F022} and DML \cite{DBLP:conf/cvpr/ZhangX23} decouple outputs for mining meaningful information. In multi-label scenarios, JointEnergy \cite{DBLP:conf/nips/WangLBL21} is the only inference technique, which jointly estimates the OOD uncertainty across all labels. While \textbf{outlier exposure} uses certain auxiliary data that disjoint with in-distribution samples for model training \cite{DBLP:conf/iclr/HendrycksMD19}. Liu \textit{et al.} \cite{DBLP:conf/nips/LiuWOL20} propose a framework for fine-tuning energy-based classifiers. Besides, some other methods like VOS \cite{DBLP:conf/iclr/DuWCL22} and POEM \cite{DBLP:conf/icml/MingFL22} adaptively sample virtual outliers to regularize decision boundary. These studies assume that feature representation forms a class-conditional Gaussian distribution. Considering the underexplored outlier exposure and the problem of JointEnergy for multi-label settings, we propose an energy-based learning framework to improve OOD detection. 

\subsection{Long-tailed OOD detection}
Traditional long-tailed learning advocates re-balancing imbalanced distributions with undersampling \cite{DBLP:conf/aaai/PengZXGHJDC19}, oversampling \cite{DBLP:conf/iccv/MullickDD19}, or re-weighting \cite{
DBLP:conf/cvpr/AlshammariWRK22, DBLP:conf/nips/WangX00CH23, DBLP:conf/icml/0001XWLHBCH24}. However, these methods proved to be inapplicable in (1) OOD detection \cite{DBLP:conf/icml/WangZZZLSW22} and (2) multi-label settings \cite{DBLP:conf/cvpr/DuarteRS21}, which could not be simply transferred to long-tailed OOD detection. Also, the research for imbalanced multi-label learning \cite{DBLP:conf/cvpr/0002021, DBLP:conf/eccv/WuH0WL20} is not well-suitable. Fortunately, Wang \textit{et al.} \cite{DBLP:conf/icml/WangZZZLSW22} propose the first learning framework to instruct OOD detection on multiclass long-tailed scenarios. Inspired by this work, this paper attempts to explore the study of long-tailed OOD detection for multi-label scenarios.

\section{Motivation}
As illustrated in Figure \ref{fig:motivation_tendency}, JointEnergy could simply discriminate the OOD sample from head ones due to the large energy gap between them. However, this success collapses when encountering the samples related to the tail classes. Due to the trivial underfitting issue on tail classes, models do not have great confidence to assign these hard samples with high logits, thereby inducing an inferior inference result. Meanwhile, \cite{DBLP:conf/icml/WangZZZLSW22} conveys that simply applying long-tailed recognition methods to OOD detection is not quite feasible. We further show that the imbalanced problem could not be solved by simply applying imbalanced multi-label learning methods. The results are shown in Table \ref{tab:multi-label methods}. Specifically, we apply three popular imbalanced multi-label recognition methods: ASL \cite{DBLP:conf/iccv/RidnikBZNFPZ21}, Focal loss \cite{DBLP:conf/iccv/LinGGHD17}, and DB Loss \cite{DBLP:conf/eccv/WuH0WL20}. Except for the BCE method, all other methods degrade the OOD detection performance. This phenomenon shows that simply applying imbalanced multi-label optimization methods is not feasible for multi-label OOD detection. Thus, this problem motivates us to seek a new OOD detection method for multi-label scenarios. In the following sections, we will present a learning framework to alleviate this issue by enlarging the energy difference between tail ID samples and OOD samples. 

\begin{table}[htbp]
  \centering
  \caption{OOD detection results using DenseNet121 with naive multi-label learning methods. All results are evaluated on the \texttt{Texture}. The best results are shown in bold.}
  \resizebox{\columnwidth}{!}{
    \begin{tabular}{c|c|ccc|c}
    \toprule
    \toprule
    Datasets & Methods & FPR95 & AUROC & AUPRC & mAP \\
    \midrule
    \multirow{4}[2]{*}{PASCAL} & DBLoss & 86.90  & 77.49  & 83.37  & 86.69  \\
          & Focal & 32.13  & 95.25  & 96.21  & 86.04  \\
          & ASL   & 32.06  & 94.90  & 95.78  & \textbf{87.65}  \\
          & BCE   & \textbf{4.50} & \textbf{98.75} & \textbf{98.69} & 87.16 \\
    \midrule
    \multirow{4}[2]{*}{COCO} & DBLoss & 99.96  & 50.09  & 83.51  & 74.11  \\
          & Focal & 89.79  & 73.21  & 95.58  & 72.54  \\
          & ASL   & 81.12  & 78.96  & 96.67  & \textbf{74.52} \\
          & BCE   & \textbf{33.81} & \textbf{95.61} & \textbf{96.77} & 74.25  \\
    \bottomrule
    \bottomrule
    \end{tabular}%
    \vspace{-10mm}
  \label{tab:multi-label methods}%
}
\end{table}%

\section{Preliminaries}
\label{sec:preliminaries}

\subsection{Problem Definition}

OOD detection could be formed as a binary classification problem. We use $\mathcal{D}_{\textup{in}}$ to denote the in-distribution and $\mathcal{D}_{\textup{out}}$ as out-of-distribution dataset. Furthermore, $\mathcal{D}_{\textup{in}}$ consists of $ \mathcal{D}_{\textup{in}}^{\textup{train}}$ and $\mathcal{D}_{\textup{in}}^{\textup{test}}$, \textit{i.e.} $\mathcal{D}_{\textup{in}} = \{ \mathcal{D}_{\textup{in}}^{\textup{train}}, \mathcal{D}_{\textup{in}}^{\textup{test}} \}$, where $\mathcal{D}_{\textup{in}}^{\textup{train}}$ refers to the training dataset and $\mathcal{D}_{\textup{in}}^{\textup{test}}$ represents the test dataset. For out-of-distribution data $\mathcal{D}_{\textup{out}}$, we term the outlier exposure samples as $\mathcal{D}_{\textup{OE}}$ and the test OOD dataset as $\mathcal{D}_{\textup{OOD}}$, respectively. In practice, two parts of distribution should meet $\mathcal{D}_{\textup{in}}\cap \mathcal{D}_{\textup{out}} = \emptyset$. The goal for the OOD detector $g_{\gamma}(\boldsymbol{x})$ is to identify whether a sample $\boldsymbol{x}$ belongs to unknown domains or not, which could be done with the following decision function:

\begin{equation}
    g_{\gamma}(\boldsymbol{x}) = \begin{cases}
    \textup{ID} & S(\boldsymbol{x}) \geq \gamma, \\ 
    \textup{OOD} & S(\boldsymbol{x}) < \gamma,
\end{cases}
\end{equation}
where $S(\boldsymbol{x})$ represents the score function and $\gamma$ refers to threshold. The sample with a high confidence score will be classified as ID and vice versa.

In multi-label OOD detection, we denote the ID input space as $\mathcal{X} \in \mathbb{R}^d$ and label space with $C$ classes as $\mathcal{Y} = \{ 0, 1 \}^{C}$ with 0 for negative and 1 for positive. Let $\mathcal{D}_{\textup{in}} = \{ \boldsymbol{x}_i, \boldsymbol{y}_i \}_{i=1}^{N}$ be the ID dataset with $N$ data points which are all sampled $i.i.d$ from the joint data distribution $\mathcal{P}_{\mathcal{X}\mathcal{Y}}$. Each label vector could be represented as $\boldsymbol{y} = [y_1, \dots, y_C]$. For OOD dataset, we define $\mathcal{D}_{\textup{out}} = \{ \boldsymbol{x}_i, y_i \}_{i=1}^{N'}$ includes $N'$ samples with only one label. Assume $\boldsymbol{z} = (\boldsymbol{x}, \boldsymbol{y}) \in \mathcal{P}_{\mathcal{X}\mathcal{Y}}$, we define $f: \mathcal{X}\rightarrow \mathcal{Y}$ as a mapping from sample input to its logit output. On top of the mapping function $f$, we define a standard pre-trained deep neural model as follows:
\begin{equation}
    f (\boldsymbol{x}; \theta) = \boldsymbol{w}_{\textup{cls}}^{\top} \cdot h(\boldsymbol{x}),
\end{equation}
where $\theta$ is the parameter of model, $\boldsymbol{w}_{\textup{cls}} \in \mathbb{R}^{d_1 \times C}$ is the classifier weight, and $h(\boldsymbol{x}) \in \mathbb{R}^{d_1}$ is the penultimate layer feature $w.r.t$ sample $\boldsymbol{x}$. For MaxLogit, $S(\boldsymbol{x}) = \max{f_{y_i}(\boldsymbol{x})}$, which only considers the maximum logit as OOD discrimination criteria. For multi-label OOD detection, JointEnergy \cite{DBLP:conf/nips/WangLBL21} is the unique energy-based score function, whose specific formula is presented as follows:

\begin{equation}
    \mathcal{E} (\boldsymbol{z}) = \sum_{i=1}^{C} \log(1 + e^{f_{y_i}(\boldsymbol{x})}),
\end{equation}
where $f_{y_i}(\boldsymbol{x})$ refers to the logit output on $i$-th class of $\boldsymbol{x}$. Unlike previous studies concerning the maximum logit, this method jointly estimates the OOD uncertainty by summing the label-wise energy across all labels, accumulating all label confidence to amplify the confidence difference between ID and OOD data effectively. However, this method would break down when encountering imbalanced multi-label data.

\section{Methodology}
\label{sec:methodology}

In this section, we will concretely introduce our proposed method EDGE in three parts. Then, a feature-based outlier selection module is presented to guide auxiliary OOD sample selection.

\subsection{Energy Distribution Gap Expansion}

The key to discriminating the OOD samples from ID ones lies in the ways of expanding the energy difference. Without knowing any open-set information, it is hard to determine a precise energy boundary for the whole feature space. Thus, we establish an effective energy-based learning framework in the multi-label setting, which consists of three main parts as follows:

\noindent \textbf{Acquire Known Distribution Information.} In order to enhance detection generalization, the first thing is to acquire useful knowledge from known domains. Here we incorporate the Binary Cross-Entropy (BCE), which is a traditional loss function in multi-label learning, as our foundational in-distribution objective, which is termed as $\mathcal{L}_{\textup{id}}$:

\begin{equation}
    \mathcal{L}_{\textup{id}} = \mathbb{E}_{\boldsymbol{z} \sim \mathcal{D}_{\textup{in}}^{\textup{train}}} \left [ \frac{1}{C} \sum_{i=1}^{C} -y_i \mathcal{L}_+(\boldsymbol{x}) - (1 - y_i) \mathcal{L}_- (\boldsymbol{x}) \right ],
    \label{eq:bce}
\end{equation}
where $\mathcal{L}_+ (\boldsymbol{x}) = \log \sigma(\boldsymbol{x})$, $\mathcal{L}_- (\boldsymbol{x}) = \log(1 - \sigma(\boldsymbol{x}))$, and $\sigma(\boldsymbol{x}) = 1 / (1 + \exp{(-f(\boldsymbol{x}; \theta))})$ is the sigmoid function \textit{w.r.t} sample $\boldsymbol{z}$.  

\noindent \textbf{Segregate Unknown Samples.} To mitigate the issue of OOD samples yielding excessively high scores, a straightforward approach is to suppress their logits across all categories. Motivated by this, we propose the confidence optimization objective $\mathcal{L}_{\textup{conf}}$ as follows:

\begin{equation}
    \mathcal{L}_{\textup{conf}} = \mathbb{E}_{\boldsymbol{z}' \sim \mathcal{D}_{\textup{OE}}} \left [ \frac{1}{C} \sum_{i=1}^{C} -\log(1 - \sigma(\boldsymbol{x}')) \right ],
    \label{eq:conf}
\end{equation}
where $\boldsymbol{z}'$ refers to an OE instance. Considering that outlier exposure data do not relate to any known class, it is reasonable for models to assign a negative label on all known classes. Consequently, we have formulated this entropy-based loss function to refine the optimization process. This objective focuses on pushing OOD samples away from the known cluster centers.

\noindent \textbf{Enlarge Distribution Distance.} The reduction in logits for OE samples allows for the accurate identification of most OOD samples. Yet, a subset of challenging cases persists among tail ID samples and semantically similar near-boundary OOD samples \cite{DBLP:journals/pami/WangXYHCH23}. Our analysis suggests that this problem stems from a critically narrow gap in energy distribution, as depicted in Figure \ref{fig:motivation_tendency} (a). The considerable overlap in distribution areas culminates in ambiguous distinctions. A type of partial contrastive learning \cite{DBLP:conf/icml/WangZZZLSW22} proves to be a potent method for distancing tail ID samples from their OOD counterparts. Nonetheless, its assumption of class-conditional independence is untenable for the unknown joint distributions encountered in multi-label settings. In response to these problems, we opt to treat the $k$ ID samples with the lowest scores collectively in contrast with all OE sample, and propose an energy-based loss function $\mathcal{L}_{\textup{gap}}$ to expand the energy gap:

\begin{equation}
    \mathcal{L}_{\textup{gap}} = \mathbb{E}_{\boldsymbol{z}' \sim \mathcal{D}_{\textup{OE}}} \left [\mathcal{E}(\boldsymbol{z}') - \mathcal{E}^{\downarrow}_{k}(\boldsymbol{z}) + m \right ]_+,
\label{eq:gap}
\end{equation}
    where $\mathcal{E}$ refers to the JointEnergy, $[a]_+ = \max\{0, a\}$, and $m$ is the margin hyper-parameter. Assume there are $n$ samples in $\mathcal{D}_{\textup{in}}^{\textup{train}}$, and we define $\mathcal{E}_{[k]}$ as top-$k$ JointEnergy scores among all ID samples \cite{DBLP:journals/pami/LyuFYH22}. Then, the average bottom-$k$ JointEnergy score of ID samples $\mathcal{E}^{\downarrow}_{k}(\boldsymbol{z})$ could be denoted as:

\begin{equation}
    \mathcal{E}^{\downarrow}_{k} (\boldsymbol{z}) = \frac{1}{k} \sum_{i=0}^{k - 1}\mathcal{E}_{[n - i]} (\boldsymbol{z}).
    \label{eq:bottomk}
\end{equation}

\begin{figure}[t]
  \centering
  \includegraphics[width=\linewidth]{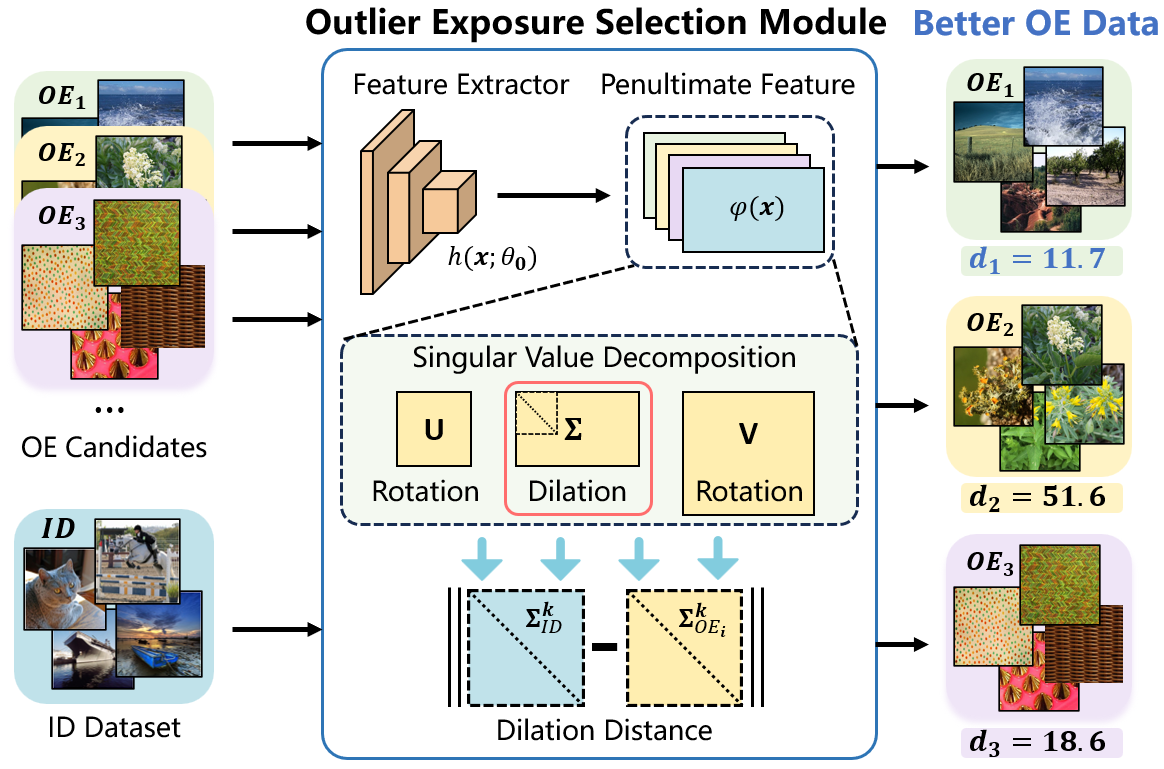}
   \caption{Illustration of Outlier Exposure Selection Module. 
}
   \label{fig:osm}
   \vspace{-4mm}
\end{figure}

\textbf{Training objective.} Benefiting the above improvements, we finally formalize our EDGE objective based on entropy-based and energy-based learning and present it as follows:

\begin{equation}
    \mathcal{L}_{\textup{edge}} = \mathcal{L}_{\textup{id}} + \alpha \cdot \mathcal{L}_{\textup{conf}} + \beta \cdot \mathcal{L}_{\textup{gap}},
    \label{eq:weg}
\end{equation}
where $\alpha, \beta$ are hyper-parameters to control weights. 

\subsection{OE Selection via Feature-based Distance}
\label{subsec:outlier selection}

It has been proved that a certain amount of auxiliary out-of-distribution data would extremely boost the discrimination of OOD detectors \cite{DBLP:conf/iclr/HendrycksMD19}. In this case, the model establishes a sufficiently strong decision boundary by previewing information from the real distribution. However, the detection effect is uncertain when using different $\mathcal{D}_{\textup{OE}}$ for training, which indicates that the success of outlier exposure is quite dependent on proper auxiliary data. Thus, a key challenge in our framework is \textit{how to select a more reliable auxiliary dataset}? By observing the feature space, it could be found the samples with similar semantic information usually assemble \cite{DBLP:conf/iclr/DuWCL22}. This trivial nature results in a serious failure to distinguish the samples with strong semantic correlation. Thus, we argue that these OOD samples could play a decisive role in improving discrimination. Based on this discovery, we propose a reasonable feature-based dilation distance to measure the suitability of outlier samples:

\begin{equation}
    d_{\textup{dilation}} = \left \| \Sigma_{\textup{in}}^k - \Sigma_{\textup{out}}^k \right \|_F,
    \label{eq:dist}
\end{equation}
where $\Sigma_{\textup{in}}^k$ and $\Sigma_{\textup{out}}^k$ are the diagonal matrix with top-$k$ singular values of ID and OOD penultimate layer feature, respectively. Motivated by current works \cite{DBLP:conf/iclr/DuWCL22, DBLP:conf/icml/MingFL22}, we think the feature representation space covers rich information, where prime singular values have a vital effect on the dilation degrees at different dimensions. Thus, an intuitive idea is that ``the closer the singular vector of two couple of samples, the more similar their feature embedding is''. Based on this finding, we could select a more informative outlier exposure dataset. To be specific, (1) we first set the batch size of ID data equal to that of OE data to ensure that the output dimensions are the same; (2) Then, solve the singular matrix separately for the two-part feature matrices; (3) Subtract the two singular matrices obtained from multiple batches, and calculate their average Frobenius norm; (4) Finally, calculate the mean of all norm results and choose the data with smaller $d_{\textup{dilation}}$ for training. We present our selection process in Figure \ref{fig:osm} and the whole training process in Algorithm \ref{alg:algorithm}.

\begin{algorithm}[tb]
\caption{Energy Distribution Gap Expansion (EDGE)}
\label{alg:algorithm}
\textbf{Input}: ID training set $\mathcal{D}_{\textup{in}}^{\textup{train}}$, OE dataset $\mathcal{D}_{\textup{OE}}$; \\
\textbf{Parameter}: model parameter $\theta$, epoch $T$, transform epoch $\tau$, hyper-parameter $\alpha, \beta, k, m$; 
\begin{algorithmic}[1] 
\STATE Compute \textit{dilation distance} between $\mathcal{D}_{\textup{in}}^{\textup{train}}$ and each individual set $\mathcal{D}_{\textup{OE}}$ by Eq. (\ref{eq:dist}). 
\FOR {$t = 0, 1, \dots, T$}
    \STATE Sample a batch of ID and OE training data from $\mathcal{D}_{\textup{in}}^{\textup{train}}$ and $\mathcal{D}_{\textup{OE}}$ respectively. 
    \IF {$t < \tau$}
        \STATE Set $\beta = 0$. 
    \ENDIF
    \STATE Calculate each loss using Eq. (\ref{eq:bce}), (\ref{eq:conf}), and (\ref{eq:gap}).
    \STATE Update parameter $\theta$ by minimizing (\ref{eq:weg}).
\ENDFOR
\end{algorithmic}
\end{algorithm}

\begin{table}[htbp]
  \centering
  \caption{Feature-based Distance between ID and OOD dataset.}
  \setlength{\tabcolsep}{3mm}{
  {\small
    \begin{tabular}{c|ccc}
    \toprule
    \multirow{2}[4]{*}{$\mathcal{D}_{\textup{out}}$} & \multicolumn{3}{c}{$\mathcal{D}_{\textup{in}}$} \\
\cmidrule{2-4}          & PASCAL & MS-COCO & NUS-WIDE \\
    \midrule
    \midrule
    Texture & 51.56  & 49.82  & 43.06  \\
    MNIST & 67.97  & 67.70  & 68.13  \\
    iSUN  & 18.60  & 22.39  & 21.00  \\
    Places50 & 14.88 & 23.05  & 17.13  \\
    LSUN  & 34.88  & 40.33  & 36.38  \\
    SVHN  & 39.24  & 42.34  & 41.28  \\
    iNaturalist & 15.50  & 20.40 & 16.20 \\
    Img-20-R & 16.90  & 23.05  & 13.60 \\
    Img-20-G & 11.75 & 9.55 & / \\
    \bottomrule
    \end{tabular}}%
  }
  \label{tab:distance}%
  \vspace{-5mm}
\end{table}%

\begin{table*}[htbp]
  \centering
  \caption{Comparison results on MS-COCO with ResNet-50. The \textcolor[rgb]{ .753,  0,  0}{\textbf{best}} and \textcolor[rgb]{ 0,  .451,  .761}{\textbf{runner-up}} performance are highlighted, and the \underline{best inference methods} are underlined. \textbf{Better result of OE methods} are bold. $\uparrow$ means the larger the better, and $\downarrow$ is opposite.}
  \resizebox{2.1\columnwidth}{!}{
  \setlength{\tabcolsep}{1.3mm}{
  {
    \begin{tabular}{c|c|ccc|ccc|ccc|c}
    \toprule
    \multicolumn{1}{c}{\multirow{2}[3]{*}{Training}} & \multirow{2}[3]{*}{Inference} & \multicolumn{3}{c|}{LSUN} & \multicolumn{3}{c|}{Texture} & \multicolumn{3}{c|}{iSUN} & \multirow{2}[3]{*}{mAP $\downarrow$} \\
\cmidrule{3-11}    \multicolumn{1}{c}{} &       & FPR95 $\downarrow$ & AUROC $\uparrow$ & AUPR $\uparrow$ & FPR95 $\downarrow$ & AUROC $\uparrow$ & AUPR $\uparrow$ & FPR95 $\downarrow$ & AUROC $\uparrow$ & AUPR $\uparrow$ &  \\
    \midrule
    \midrule
    \multirow{10}[2]{*}{BCE} & MaxLogit & 1.08  & 99.63  & 99.90  & 11.19  & 96.97  & 99.44  & 29.80  & 89.93  & 96.02  & \multirow{10}[2]{*}{73.79} \\
          & MSP   & 22.41  & 95.13  & 98.65  & 45.51  & 87.64  & 97.62  & 60.58  & 79.99  & 92.89  &  \\
          & ODIN  & 1.31  & 99.23  & 99.81  & 11.51  & 96.31  & 99.28  & 33.42  & 88.70  & 95.71  &  \\
          & Mahalanobis & 10.18  & 98.02  & 99.54  & 18.21  & 96.28  & 99.42  & 97.78  & 63.25  & 88.26  &  \\
          & Energy Score & 0.75  & 99.72  & 99.92  & 9.11  & 97.33  & 99.49  & 27.51  & 90.28  & 96.16  &  \\
          & ReAct & 3.38  & 98.98  & 99.69  & 8.58  & 97.78  & 99.62  & 27.39  & 89.88  & 95.60  &  \\
          & ViM   & \underline{0.27}  & \underline{99.91}  & \underline{99.98}  & \underline{5.76}  & \underline{98.31}  & \underline{99.71}  & 36.41  & 89.01  & 96.04  &  \\
          & ASH   & 1.38  & 98.88  & 98.90  & 8.24  & 97.82  & 99.61  & 30.45  & 90.21  & 96.81  &  \\
          & MaxEnergy & 1.08  & 99.63  & 99.90  & 11.19  & 96.97  & 99.44  & 29.80  & 89.83  & 96.02  &  \\
          & JointEnergy & 0.48  & 99.80  & 99.95  & 7.20  & 98.15  & 99.71  & \underline{26.05}  & \underline{92.04}  & \underline{97.47}  &  \\
    \midrule
    \multirow{2}[1]{*}{OE} & MaxEnergy & \textbf{5.79}  & \textbf{98.78}  & \textbf{99.72}  & \textbf{9.38}  & \textbf{98.09}  & \textbf{99.72}  & \textbf{25.42}  & \textbf{93.41}  & \textbf{97.95}  & \multirow{2}[1]{*}{68.47} \\
          & JointEnergy & 69.19  & 65.80  & 81.33  & 60.30  & 71.41  & 89.58  & 57.67  & 78.84  & 88.62  &  \\
    \multirow{2}[1]{*}{EnergyOE} & MaxEnergy & 18.01  & 97.28  & 99.34  & 27.85  & 92.92  & 98.69  & 49.29  & 86.36  & 95.99  & \multirow{2}[1]{*}{72.37} \\
          & JointEnergy & 33.51  & 81.38  & 88.01  & 38.60  & 80.49  & 92.37  & 46.29  & 81.52  & 89.35  &  \\
    \midrule
    \multirow{2}[2]{*}{EDGE} & MaxEnergy & \textcolor[rgb]{ 0,  .451,  .761}{\textbf{0.07}} & \textcolor[rgb]{ .753,  0,  0}{\textbf{99.95}} & \textcolor[rgb]{ .753,  0,  0}{\textbf{99.99}} & \textcolor[rgb]{ 0,  .451,  .761}{\textbf{4.18}} & \textcolor[rgb]{ 0,  .451,  .761}{\textbf{98.90}} & \textcolor[rgb]{ 0,  .451,  .761}{\textbf{99.82}} & \textcolor[rgb]{ 0,  .451,  .761}{\textbf{20.15}} & \textcolor[rgb]{ 0,  .451,  .761}{\textbf{93.90}} & \textcolor[rgb]{ 0,  .451,  .761}{\textbf{97.84}} & \multirow{2}[2]{*}{73.50} \\
          & JointEnergy & \textcolor[rgb]{ .753,  0,  0}{\textbf{0.01}} & \textcolor[rgb]{ 0,  .451,  .761}{\textbf{99.91}} & \textcolor[rgb]{ 0,  .451,  .761}{\textbf{99.98}} & \textcolor[rgb]{ .753,  0,  0}{\textbf{3.09}} & \textcolor[rgb]{ .753,  0,  0}{\textbf{99.17}} & \textcolor[rgb]{ .753,  0,  0}{\textbf{99.88}} & \textcolor[rgb]{ .753,  0,  0}{\textbf{18.45}} & \textcolor[rgb]{ .753,  0,  0}{\textbf{94.86}} & \textcolor[rgb]{ .753,  0,  0}{\textbf{98.40}} &  \\
    \bottomrule
    \end{tabular}}}}
  \label{tab:overall_results_coco}%
  \vspace{-4mm}
\end{table*}%

\section{Experiments}
\label{sec:experiments}

In this section, we present and analyze the overall OOD detection performance of EDGE framework. \textbf{Due to the space limitation, we present partial vital details and results in the main paper and put others in Appendix.} 

\subsection{Experimental settings}

\textbf{In-distribution Datasets.} We adopt three real-world benchmark multi-label image annotation datasets as our training sets, \textit{i.e.}, PASCAL-VOC \cite{DBLP:journals/ijcv/EveringhamEGWWZ15}, MS-COCO \cite{DBLP:conf/eccv/LinMBHPRDZ14} and NUS-WIDE \cite{DBLP:conf/civr/ChuaTHLLZ09}. PASCAL-VOC consists of 22,531 images with 20 categories; MS-COCO contains 122,218 annotation images with 80 categories, where 82,783 images are for training, 40,504 for validation, and 40,775 for testing; and NUS-WIDE covers 269,648 real-world images with 81 categories, where we reserve 119,986 images of the training set and 80,283 images of the test set, respectively.

\noindent \textbf{Out-of-distribution Datasets.} To fully evaluate the robustness of the model on out-of-distribution examples, we follow previous literature and choose the following benchmark OOD datasets as main experimental candidates: (a subset of) \texttt{ImageNet-22K} \cite{DBLP:conf/cvpr/DengDSLL009}, \texttt{Textures} \cite{DBLP:conf/cvpr/CimpoiMKMV14}, \texttt{Places50} \cite{DBLP:journals/pami/ZhouLKO018}, \texttt{iSUN} \cite{DBLP:journals/corr/XuEZFKX15}, \texttt{iNaturalist} \cite{DBLP:conf/cvpr/HornASCSSAPB18}, \texttt{LSUN-C} \cite{DBLP:journals/corr/YuZSSX15}, and \texttt{SVHN} \cite{2011Netzer}, where the categories of all samples do not overlap with those in used ID datasets and ImageNet-1K \cite{DBLP:conf/cvpr/DengDSLL009}. Owing to the outlier exposure selection module, we present the distance between each ID and OOD dataset in Table \ref{tab:distance}, and select \texttt{Img-20-G} as $\mathcal{D}_{\textup{OE}}$. Note that NUS-WIDE overlaps some classes of \texttt{Img-20-G}, we thus follow \cite{DBLP:conf/nips/WangLBL21} and adopt another different subset \texttt{Img-20-R}. More details about OOD datasets can be found in Appendix. 

\noindent \textbf{Implementation details.} All the experiments are run on NVIDIA
GeForce RTX 3090 implemented by PyTorch and conducted on the ResNet-50 \cite{DBLP:conf/cvpr/HeZRS16} and VGG16 \cite{DBLP:journals/corr/SimonyanZ14a} pre-trained on ImageNet-1K \cite{DBLP:conf/cvpr/DengDSLL009}. We randomly horizontally flip all input images and resize them to 256 $\times$ 256. We use stochastic gradient descent (SGD) \cite{DBLP:conf/icml/SutskeverMDH13} with a momentum of 0.9 and a weight decay of $1 \times 10^{-4}$. For trivial BCE models, we use the Adam \cite{DBLP:journals/corr/KingmaB14} to optimize. The main hyper-parameters $\alpha$, $\beta$, and $m$ take value from $\{1e+1, 1, 1e-1, 1e-2, 1e-3 \}$, $\{ 1, 1e-1, 1e-2, 1e-3, 1e-4 \}$, and $\{ 0, 1, 2, 3, 4, 5 \}$.

\begin{table*}[t]
  \centering
  \caption{Ablative results on different combinations of loss functions, where \checkmark means the utilization of the corresponding loss term. The \textcolor[rgb]{ .753,  0,  0}{\textbf{best}} and \textcolor[rgb]{ 0,  .451,  .761}{\textbf{runner-up}} OOD performance is highlighted.}
  \setlength{\tabcolsep}{1mm}{
  {\small
    \begin{tabular}{c|ccc|ccc|ccc|ccc|c}
    \toprule
    \multirow{2}[4]{*}{$\mathcal{D}_{\textup{in}}$} & \multirow{2}[4]{*}{$\mathcal{L}_{\textup{bce}}$} & \multirow{2}[4]{*}{$\mathcal{L}_{\textup{conf}}$} & \multirow{2}[4]{*}{$\mathcal{L}_{\textup{gap}}$} & \multicolumn{3}{c|}{Places50} & \multicolumn{3}{c|}{Texture} & \multicolumn{3}{c|}{iSUN} & \multirow{2}[4]{*}{mAP $\uparrow$} \\
\cmidrule{5-13}          &       &       &       & FPR95 $\downarrow$ & AUROC $\uparrow$ & AUPR $\uparrow$ & FPR95 $\downarrow$ & AUROC $\uparrow$ & AUPR $\uparrow$ & FPR95 $\downarrow$ & AUROC $\uparrow$ & AUPR $\uparrow$ &  \\
    \midrule
    \midrule
    \multirow{4}[2]{*}{MS-COCO} & \checkmark &       &       & 0.48  & 99.80  & 99.95  & 7.20  & 98.15  & 99.71  & 26.05  & 92.04  & 97.46  & 73.79  \\
          & \checkmark & \checkmark &       & \textcolor[rgb]{ 0,  .451,  .761}{\textbf{0.02}} & 99.90  & 99.95  & \textcolor[rgb]{ 0,  .451,  .761}{\textbf{5.38}} & \textcolor[rgb]{ 0,  .451,  .761}{\textbf{98.69}} & \textcolor[rgb]{ 0,  .451,  .761}{\textbf{99.77}} & \textcolor[rgb]{ 0,  .451,  .761}{\textbf{20.39}} & \textcolor[rgb]{ 0,  .451,  .761}{\textbf{93.03}} & \textcolor[rgb]{ 0,  .451,  .761}{\textbf{97.54}} & 68.18  \\
          & \checkmark &       & \checkmark & 0.01  & \textcolor[rgb]{ 0,  .451,  .761}{\textbf{99.93}} & \textcolor[rgb]{ 0,  .451,  .761}{\textbf{99.98}} & 8.46  & 97.73  & 99.61  & 43.14  & 86.93  & 95.61  & 57.98  \\
          & \checkmark & \checkmark & \checkmark & \textcolor[rgb]{ .753,  0,  0}{\textbf{0.01}} & \textcolor[rgb]{ .753,  0,  0}{\textbf{99.93}} & \textcolor[rgb]{ .753,  0,  0}{\textbf{99.98}} & \textcolor[rgb]{ .753,  0,  0}{\textbf{3.09}} & \textcolor[rgb]{ .753,  0,  0}{\textbf{99.17}} & \textcolor[rgb]{ .753,  0,  0}{\textbf{99.88}} & \textcolor[rgb]{ .753,  0,  0}{\textbf{18.45}} & \textcolor[rgb]{ .753,  0,  0}{\textbf{94.86}} & \textcolor[rgb]{ .753,  0,  0}{\textbf{98.40}} & 73.50  \\
    \bottomrule
    \end{tabular}}%
   }
  \label{tab:ablation}%
  \vspace{-3mm}
\end{table*}%

\begin{table*}[htbp]
  \centering
  \caption{FPR95 (F$\downarrow$) and AUROC (AR$\uparrow$) Performance of OOD detection and ID classification when using different datasets as $\mathcal{D}_{\textup{OE}}$ and $\mathcal{D}_{\textup{OOD}}$ on NUS-WIDE. The \textbf{best results} on the same $\mathcal{D}_{\textup{OOD}}$ are bold.}
  \renewcommand\arraystretch{0.85}
  \setlength{\tabcolsep}{2mm}{
    \begin{tabular}{c|c|c|cccccccc}
    \toprule
    $\mathcal{D}_{\textup{OE}}$ & $d_{\textup{dilation}} \downarrow$ & $\mathcal{D}_{\textup{OOD}}$ & Texture & iSUN  & Img-20-R & Places50 & iNaturalist & MNIST & SVHN & Average \\
    \midrule
    \midrule
    Texture & 43.06  &       & /     & 70.92  & 30.00  & 67.83  & 34.31  & 9.70  & 0.20  & 35.49 \\
    iSUN  & 21.00  & \multicolumn{1}{c|}{F $\downarrow$} & 10.62  & /     & \textbf{15.73} & \textbf{5.01} & 8.28  & 5.78  & 2.25 & 7.95 \\
    Img-20-R & \textbf{13.61} &       & \textbf{2.46} & \textbf{9.10} & /     & 26.36  & \textbf{0.79} & \textbf{0.00} & \textbf{0.00} & \textbf{6.45} \\
    \midrule
    Texture & 43.06  &       & /     & 67.65  & 92.47  & 69.20  & 90.74  & 97.23  & 99.65 & 86.16 \\
    iSUN  & 21.00  & \multicolumn{1}{l|}{AR $\uparrow$} & 97.16  & /     & \textbf{97.03} & \textbf{94.40} & 98.02  & 97.69  & 99.02 & 97.21 \\
    Img-20-R & \textbf{13.61} &       & \textbf{99.34} & \textbf{96.92} & /     & 89.35  & \textbf{99.83} & \textbf{99.85} & \textbf{99.91} & \textbf{97.53} \\
    \bottomrule
    \end{tabular}}%
  \label{tab:outiler_test_nus}%
  \vspace{-3mm}
\end{table*}%

\noindent \textbf{Competitors.} Totally, we selected 10 inference-based approaches as comparisons, including MaxLogit \cite{DBLP:conf/icml/HendrycksBMZKMS22}, MSP \cite{DBLP:conf/iclr/HendrycksG17}, ODIN \cite{DBLP:conf/iclr/LiangLS18}, Energy \cite{DBLP:conf/nips/LiuWOL20}, Mahalanobis \cite{DBLP:conf/nips/LeeLLS18}, ReAct \cite{DBLP:conf/nips/SunGL21}, ViM \cite{DBLP:conf/cvpr/Wang0F022}, ASH \cite{DBLP:conf/iclr/DjurisicBAL23}, MaxEnergy, and JointEnergy \cite{DBLP:conf/nips/WangLBL21}. Note that all results obtained by inference methods are based on the model optimizing with BCE loss. We adopt OE \cite{DBLP:conf/iclr/HendrycksMD19} and EnergyOE \cite{DBLP:conf/nips/LiuWOL20} as outlier exposure comparisons.

\begin{figure}[htbp]
    \centering
    \begin{subfigure}{0.49\linewidth}
        \includegraphics[width=1.0\linewidth]{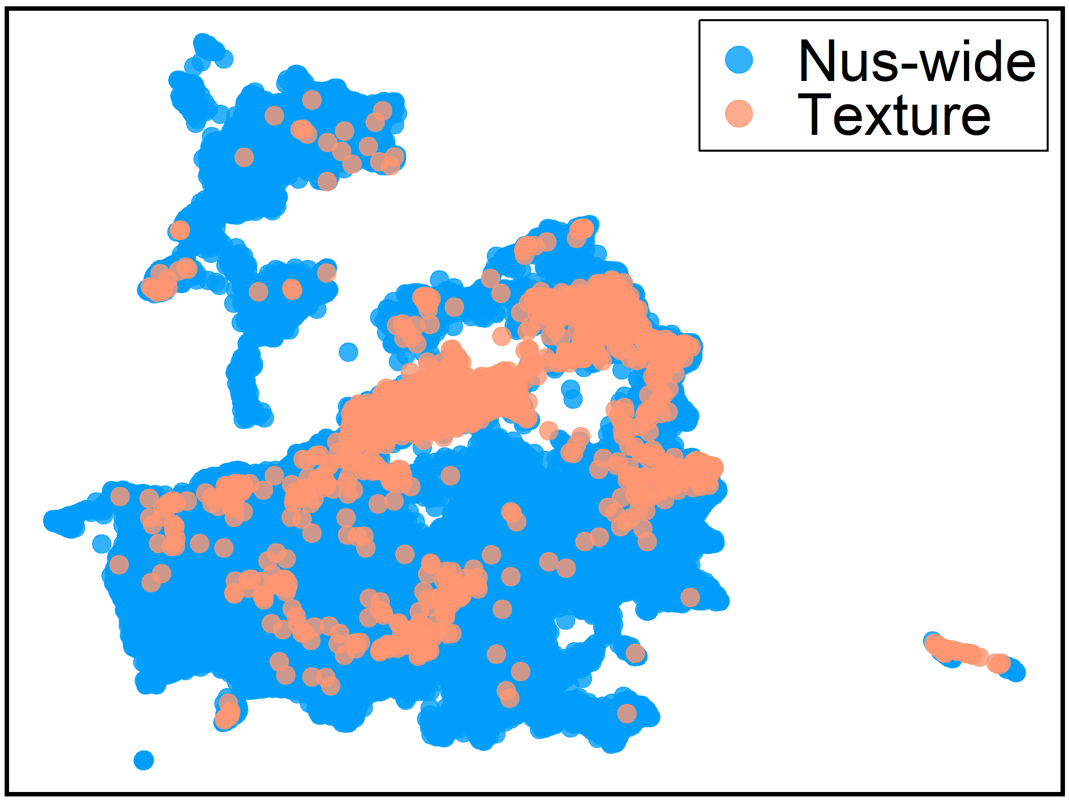}
        \caption{w/o EDGE, Texture}
        \label{fig:BCE_places}
    \end{subfigure}
    \hfill
    \begin{subfigure}{0.49\linewidth}
        \includegraphics[width=1.0\linewidth]{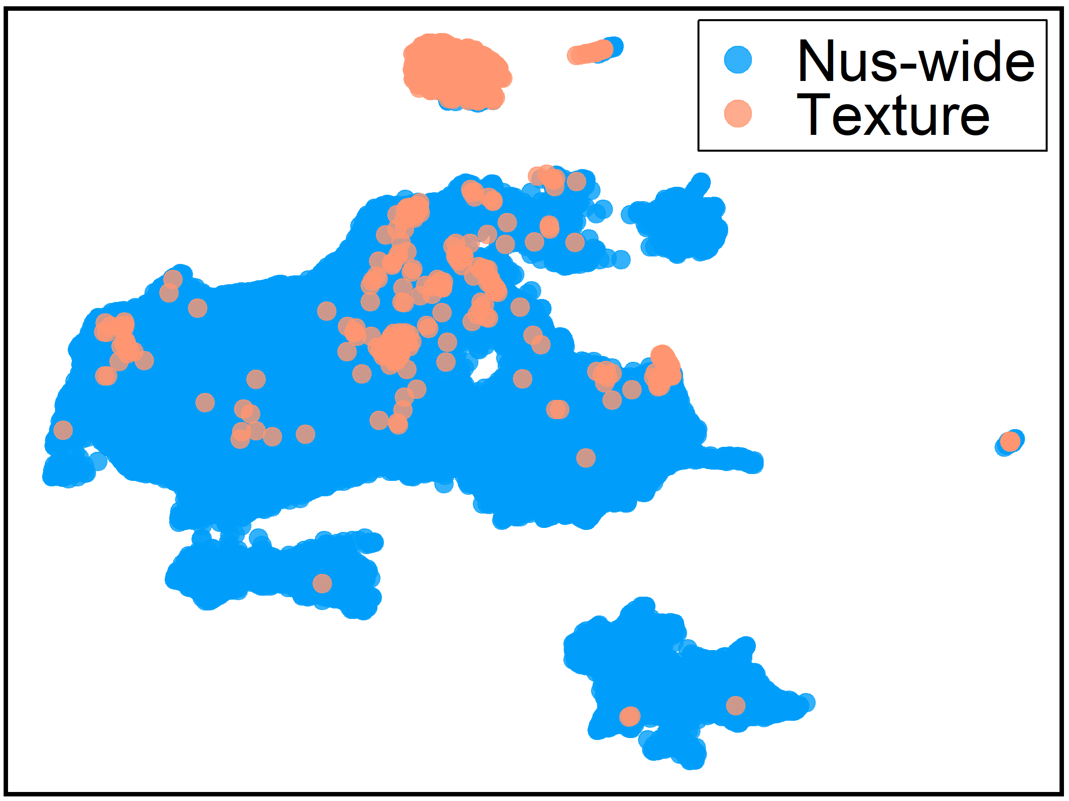}
        \caption{w/ EDGE, Texture}
        \label{fig:EDGE_places}
    \end{subfigure}
    \vspace{-0.2cm}
    \caption{UMAP visualization of feature embedding results.}
    \label{fig:umap}
    \vspace{-6mm}
\end{figure}

\subsection{Main Results}
\textbf{Overall Performance.} We present the overall comparative results on various ID and OOD datasets in Table \ref{tab:overall_results_coco}. We test various inference methods as well as different outlier exposure methods under the same training criterion and pre-trained model. We get some observations below: (1) Our proposed EDGE significantly outperforms all comparative methods; (2) Our EDGE achieves a significant improvement in OOD detection with a less loss of ID performance, promising a relatively balanced result between known and unknown domains;  (3) The complexity of the $\mathcal{D}_{\textup{in}}$ and $\mathcal{D}_{\textup{OE}}$ directly affects both ID classification and OOD detection. In larger-scale NUS-WIDE, most comparative methods produce serious degeneration, especially on \texttt{iSUN}, where this dataset contains some similar scene images, like the jungle, with the images in NUS-WIDE. However, our method still maintains promising performance while resisting hard samples; (4) An interesting point is that both OE and EnergyOE achieve an unstable performance, even poorer than some post-hoc methods. It indicates that simply transferring general outlier exposure methods is not always feasible; (5) OE and EnergyOE perform better when utilizing MaxEnergy while not JointEnergy. A reasonable explanation is that traditional methods are applied in multi-class scenarios and prefer to identify OOD samples with their maximum logit. This nature makes them function poorly in multi-label scenarios. 

\noindent \textbf{Visualization.} To further show the expansion effect, we visualize the results of feature embedding output in Figure \ref{fig:umap}. The blue and orange parts represent the cluster of ID data and OOD data, respectively. We qualitatively analyze the discriminative effect when model training with/without our EDGE framework. Compared with general models, the EDGE-based model could accurately separate OOD samples from ID data and cluster them. This demonstrates the superiority of our proposed method.

\begin{figure}[htbp]
    \centering
    \begin{subfigure}{0.49\linewidth}
        \includegraphics[width=1.0\linewidth]{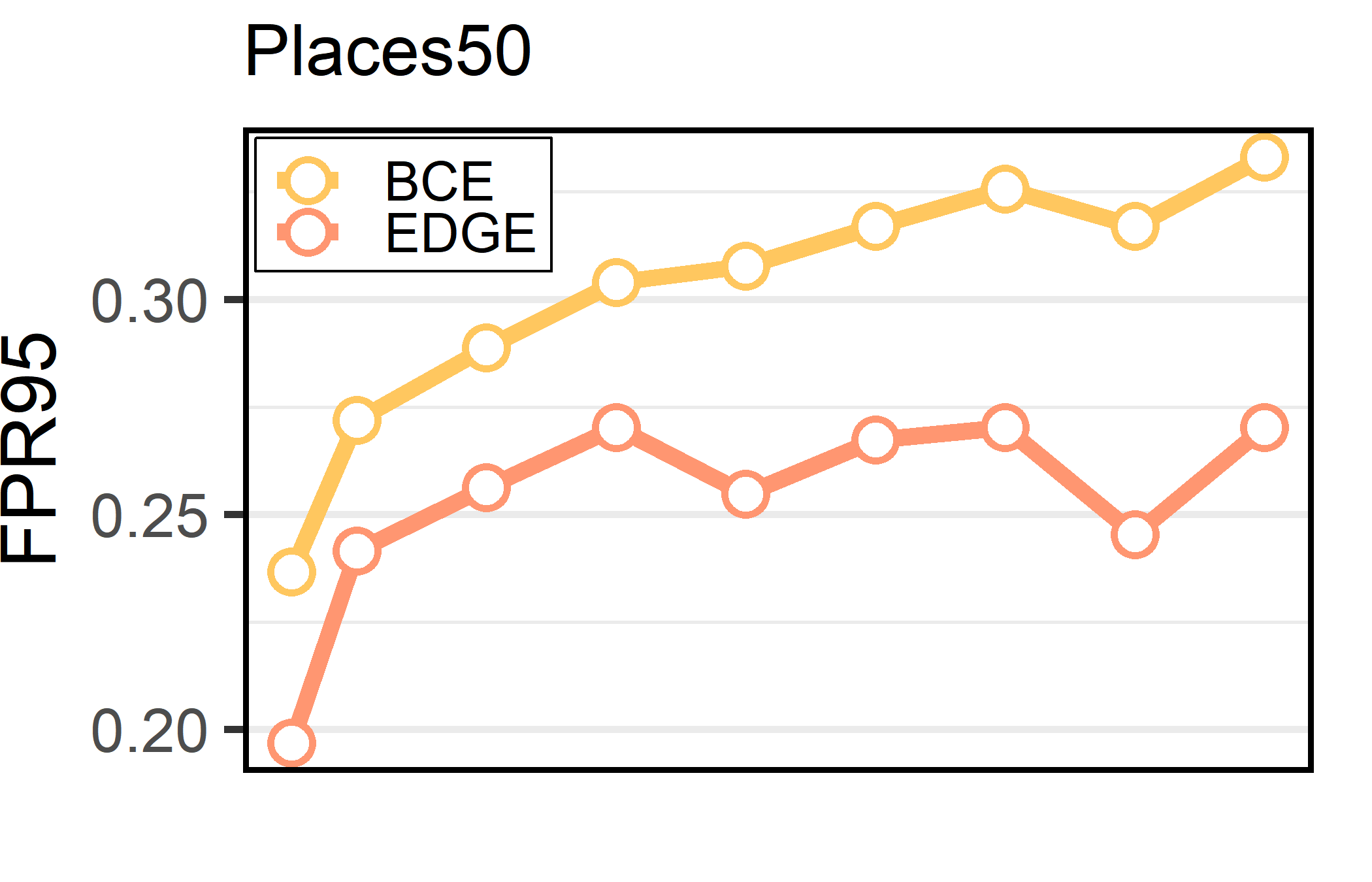}
        \label{fig:Places50_fpr}
    \end{subfigure}
    \hfill
    \vspace{-0.7cm}
    \begin{subfigure}{0.49\linewidth}
        \includegraphics[width=1.0\linewidth]{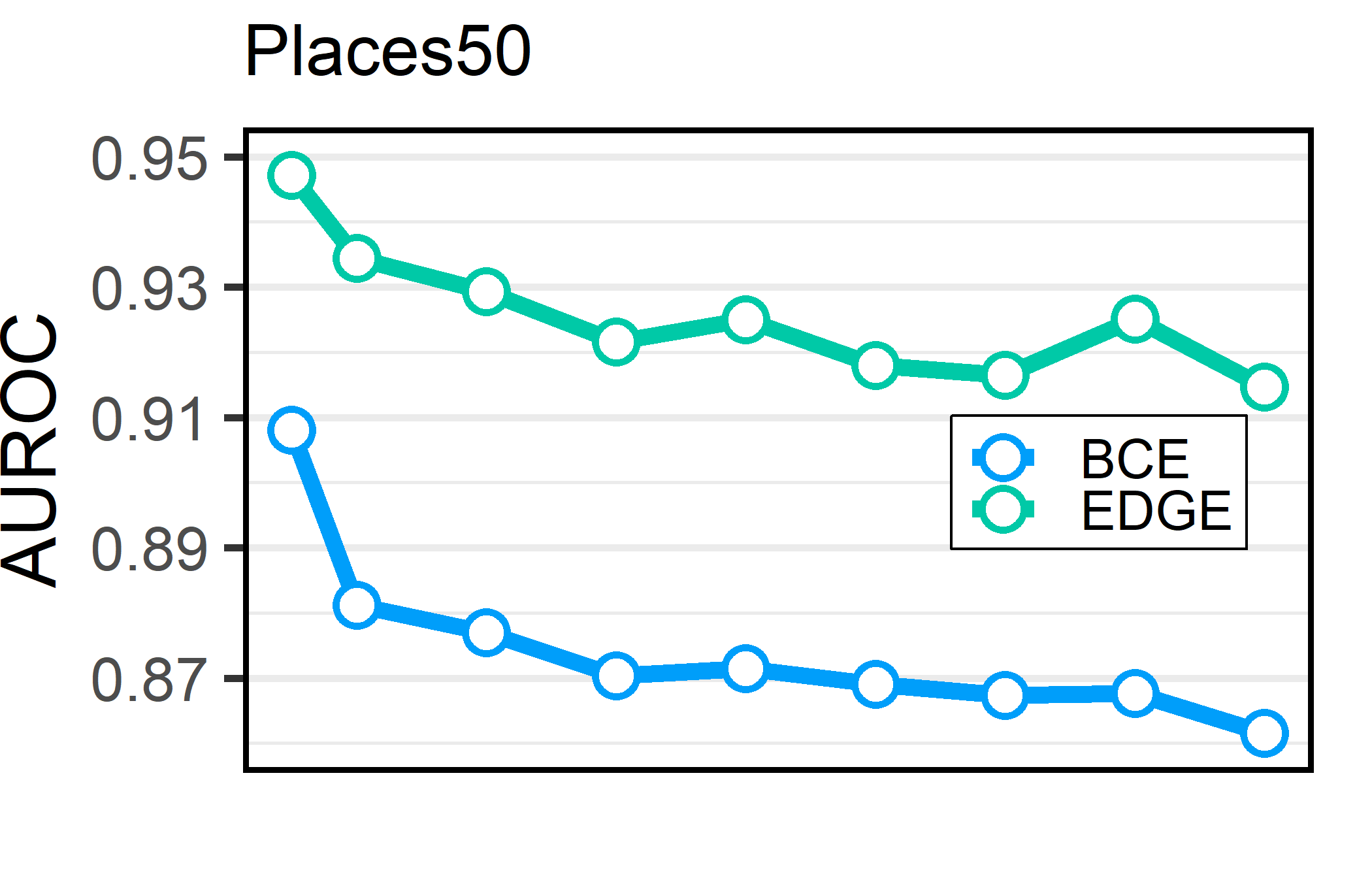}
        \label{fig:Places50_auc}
    \end{subfigure}
    \vspace{-0.7cm}
    \begin{subfigure}{0.49\linewidth}
        \includegraphics[width=1.0\linewidth]{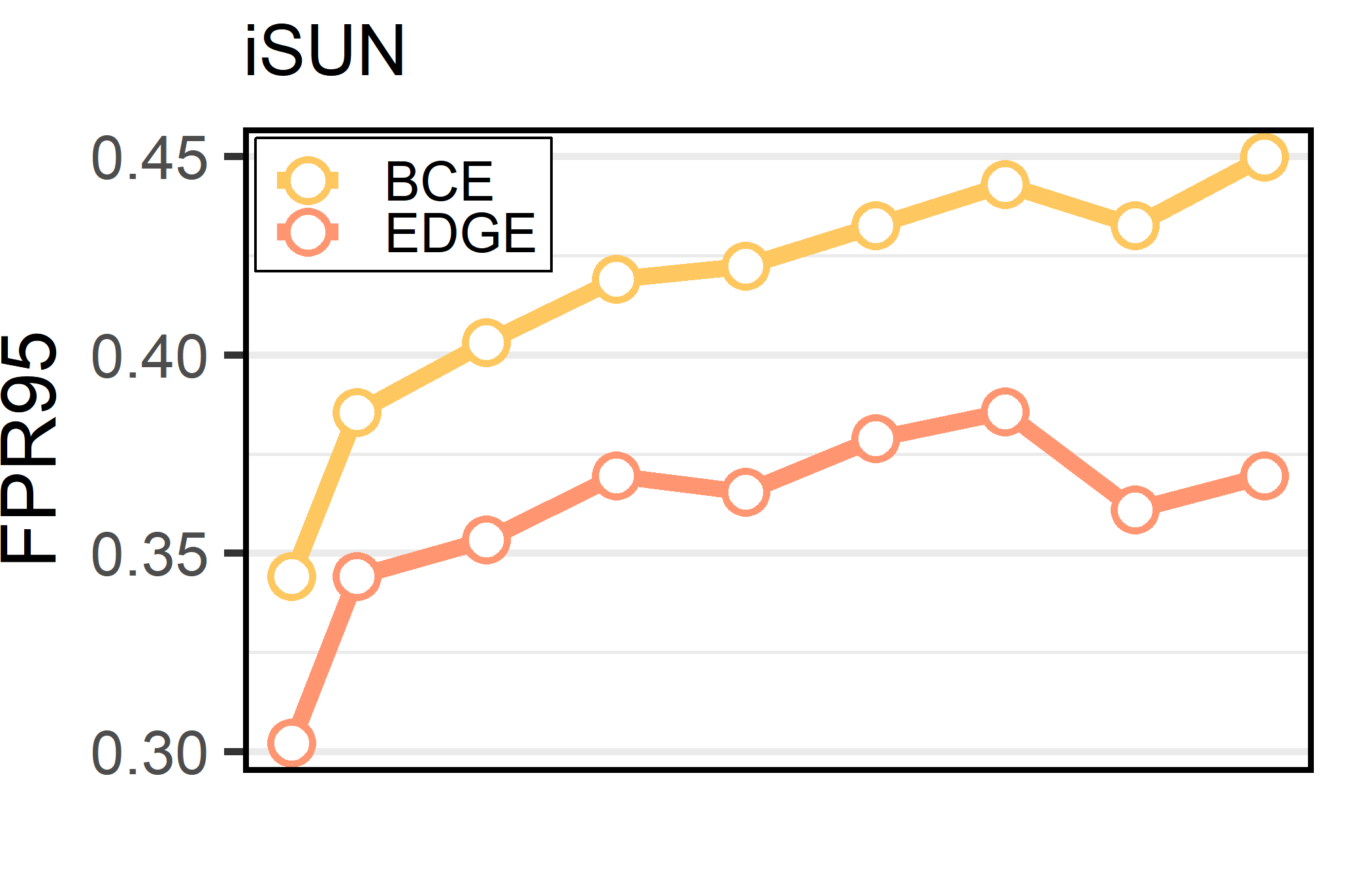}
        \label{fig:iSUN_fpr}
    \end{subfigure}
    \hfill
    \begin{subfigure}{0.49\linewidth}
        \includegraphics[width=1.0\linewidth]{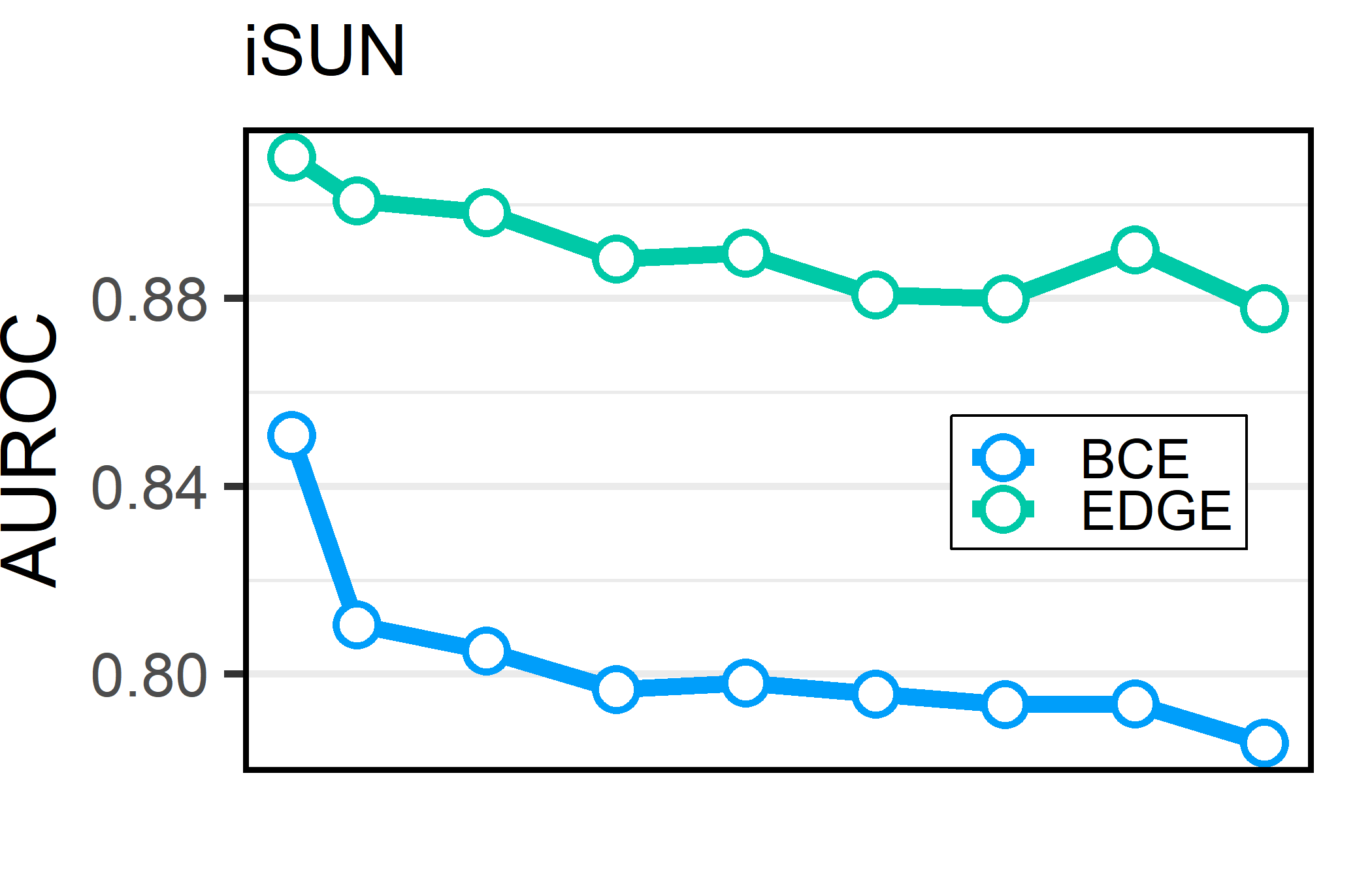}
        \label{fig:iSUN_auc}
    \end{subfigure}
    \caption{Performance change curve on PASCAL-VOC. \textit{The closer to the right, the larger the proportion of tail samples.}}
    \label{fig:tendency_pascal}
    \vspace{-6mm}
\end{figure}

\noindent \textbf{Performance on tail samples.} Figure \ref{fig:tendency_pascal} shows several groups of comparative results on tail samples on PASCAL-VOC. On top of the overall improvement, our method achieves significant tail improvement. As the proportion of tail samples increases, the performance of EDGE is less negatively impacted. These results indicate the solving of the problem that is illustrated in the motivation.

\noindent \textbf{Effectiveness of OE Selection.} We record the performance of OOD detection based on different outlier exposure data in Table \ref{tab:outiler_test_nus}. The overall OOD performance on \texttt{Img-20-R} is obviously better than others, where we observe it also holds a relatively small $d_{\textup{dilation}}$. Moreover, for any two datasets, selecting the one with a smaller $d_{\textup{dilation}}$ as $\mathcal{D}_{\textup{OE}}$ could achieve better performance. These intuitive results are consistent with our elaboration in the previous section and validate the effectiveness of the selection method. 

\noindent \textbf{Sensitivity Analysis.} We conducted a series of hyperparameter sensitivity analyses on PASCAL-VOC. Due to space constraints, we include all results in Appendix.

\subsection{Ablation studies}

To further test the impact of each part in EDGE framework, we carry out the ablation study on three benchmarks and report the results in Table \ref{tab:ablation}. From this table, we could obtain the following observations: (1) Whichever of $\mathcal{L}_{\textup{conf}}$ and $\mathcal{L}_{\textup{gap}}$ is combined with $\mathcal{L}_{\textup{id}}$ could lead to a significant improvement, demonstrating the improvement of increments. (2) While $\mathcal{L}_{\textup{conf}}$ and $\mathcal{L}_{\textup{gap}}$ take turns in the lead, the former has a larger effect on keeping the performance of ID classification, indicating that optimizing the logit information plays a dominant role. (3) The combination of all loss terms could achieve not only a relatively comprehensive improvement in OOD detection, but also a better ID classification performance compared to other OE methods. In general, each module explicitly displays a certain positive effect.

\vspace{-0.3cm}

\section{Conclusion}
\label{sec:conclusion}

In this paper, we discover the shortcomings in multi-label OOD detection, \textit{i.e.}, a serious long-tailed phenomenon caused by trivial optimization methods. To fix this issue, we introduce outlier exposure into multi-label classification. On top of this, we propose a simple yet effective learning framework dubbed EDGE. This method first separately optimizes ID samples and OE samples to establish a reliable performance of ID classification. It then expands the energy gap between ID and OOD samples by an energy-boundary learning strategy. Besides, a feature-based selection module is devised to boost the performance of model learning. Finally, extensive experiments consistently validate the effectiveness of the proposed method on both ID classification and OOD detection. 

\section{Acknowledgments}

This work was supported in part by the National Key R\&D Program of China under Grant 2018AAA0102000, in part by National Natural Science Foundation of China: 62236008, U21B2038, U23B2051, 61931008, 62122075, 62206264, and 92370102, in part by Youth Innovation Promotion Association CAS, in part by the Strategic Priority Research Program of the Chinese Academy of Sciences, Grant No. XDB0680000, in part by the Innovation Funding of ICT, CAS under Grant No.E000000, in part by the China National Postdoctoral Program for Innovative Talents under Grant BX20240384.

\bibliography{aaai25}

\end{document}